\documentclass[supp]{new_tlp} 
\usepackage{times,helvet,courier}
\usepackage{latexsym}
\usepackage{amssymb}
\usepackage{xspace}
\usepackage{url}
\usepackage{graphicx}
\graphicspath{{./figures/}}
\usepackage{epsfig}
\usepackage{amsmath}
\usepackage{subfigure}
\usepackage{lscape}
\usepackage{paralist}

\usepackage{rotating}
\usepackage{bigstrut}
\usepackage{multirow}
\usepackage{booktabs}
\usepackage{float}
\floatstyle{boxed}
\restylefloat{figure}

\usepackage{supp}
\pagerange{\pageref{firstpage}--\pageref{lastpage}}
\pubyear{2013}
\pubauthor{Calimeri}

\newif\ifdotikz\dotikzfalse

\ifdotikz
\usepackage{tikz}
\usetikzlibrary{positioning}
\pgfrealjobname{lpnmr13}
\usetikzlibrary{shapes,arrows,backgrounds,chains,%
matrix,patterns,arrows,decorations.pathmorphing,decorations.pathreplacing,%
positioning,fit,calc,decorations.text,shadows%
}
\else
\long\def\beginpgfgraphicnamed#1#2\endpgfgraphicnamed{\includegraphics{#1}}
\fi

\usepackage{framed}
\usepackage{amssymb}
\usepackage{latexsym}

\newcommand\nameFont[1]{{\fontencoding{OT1}\fontfamily{cmss}\selectfont{#1}}}
\def\nf{\nameFont}

\def\aspcoretwo{\nf{ASP-Core-2}\xspace}

\def\team{Model\&Solve\xspace}
\def\solver{System\xspace}

\newcommand\nop[1]{}

\newcommand{\countagg}{\ensuremath{\mathtt{\# count}}}
\newcommand{\sumagg}{\ensuremath{\mathtt{\# sum}}}
\newcommand{\minagg}{\ensuremath{\mathtt{\# min}}}
\newcommand{\maxagg}{\ensuremath{\mathtt{\# max}}}

\newenvironment{indentnew}[1]%
{\begin{list}{}%
         {\setlength{\leftmargin}{#1}}%
         \item[]%
} {\end{list}}

\newcommand{\gap}{\\[-3mm]}
\newcommand{\problem}[1]{\textit{#1}}
\newcommand{\bas}{$^\ast$}

\newcommand{\Dec}{D}
\newcommand{\Opt}{O}
\newcommand{\Que}{Q}

\mathchardef\hn="2D
\newcommand{\pr}{\ensuremath{\mathtt{non\hn{}tight}}}
\newcommand{\ch}{\ensuremath{\mathtt{choice}}}
\newcommand{\chb}{\ensuremath{^{\raisebox{-1pt}{$\mathtt{\scriptscriptstyle\#}$}}}}
\newcommand{\chc}{\ch\chb}
\newcommand{\ag}{\ensuremath{\mathtt{aggr}}}
\newcommand{\di}{\ensuremath{\mathtt{disj}}}
\newcommand{\lev}{\ensuremath{\mathtt{level}}}
\newcommand{\out}{n/a}
\newcommand{\ba}{\ensuremath{\mathtt{basic}}}

\def\clasp{{\sc {\small clasp}}} 
\def\claspmt{{\sc {\small clasp-mt}}} 

\def\Gringof{{\sc {\small gringo-4}}} 

\def\wasp{{\sc {\small wasp}}} 

  {\end{list}}

\title[The Design of the Fifth Answer Set Programming Competition]{%
The Design of the Fifth Answer Set Programming Competition%
}

\author[Francesco Calimeri, Martin Gebser, Marco Maratea, Francesco Ricca]{%
Francesco Calimeri and Francesco Ricca
\\ Dipartimento di Matematica e Informatica, Universit{\`a} della Calabria, Italy
\and Martin Gebser
\thanks{Also affiliated with the University of Potsdam, Germany.}
\\ Helsinki Institute for Information Technology HIIT
\\ Department of Information and Computer Science, School of Science, Aalto University, Finland
\and Marco Maratea
\\ Dipartimento di Informatica, Bioingegneria, Robotica e Ingegneria dei Sistemi, Universit{\`a} di Genova, Italy
}

\begin{document}

\maketitle

\begin{abstract}
Answer Set Programming (ASP) is a well-established paradigm of declarative programming that has been developed in the field of logic programming and nonmonotonic reasoning. Advances in ASP solving technology are customarily assessed in competition events, as it happens for other closely-related problem-solving technologies like    SAT/SMT, QBF, Planning and Scheduling.
ASP Competitions are (usually) biennial events; however, the Fifth ASP Competition departs from tradition, in order to join the FLoC Olympic Games at the Vienna Summer of Logic 2014, which is expected to be the largest event in the history of logic.
This edition of the ASP Competition series is jointly organized by the University of Calabria (Italy), the Aalto University (Finland), and the University of Genova (Italy), and is affiliated with the 30th International Conference on Logic Programming (ICLP 2014).
It features a completely re-designed setup, with novelties involving the design of tracks, the scoring schema, and the adherence to a fixed modeling language in order to push the adoption of the \aspcoretwo standard. Benchmark domains are taken from past editions, and best system packages submitted in 2013 are compared with new versions and solvers.
\end{abstract}

\section{Introduction}\label{sec:introduction}

Answer Set Programming~\cite{bara-2002,%
 eite-etal-2000c,%
 eite-etal-97f,%
 eite-etal-2009-primer,%
 gelf-leon-02,%
 gelf-lifs-91,lifs-99a,%
 mare-trus-99} is a well-established 
declarative programming approach to knowledge
 representation and reasoning, proposed in the area of nonmonotonic
 reasoning and logic programming, with close relationship to other
formalisms such as SAT, SAT Modulo Theories, Constraint Handling
Rules, PDDL, and many others. The Answer Set Programming (ASP) Competition is a (usually) biennial event whose goal is to access the
state of the art in ASP solving (see, e.g., \cite{wasp-lpnmr-13,%
 dalp-etal-2009-fi,%
 gekasc13a,%
 gks2012-aij,%
 glm2006-jar,%
 janh-etal-2005-tocl,%
 leon-etal-2002-dlv,%
 lin-zhao-2004,%
 ljn-kr-12,%
 maar-etal-2008-minisatid,%
 simo-etal-2002} %
on challenging benchmarks. This year we depart from the usual
timeline, and the event takes place one year after the Fourth ASP Competition\footnote{\url{https://www.mat.unical.it/aspcomp2013/}}; basically, the two main reasons are: $(i)$ to be part of the Vienna Summer of Logic (VSL)\footnote{\url{http://vsl2014.at/}}, which is expected to be the largest event in the history of logic, and $(ii)$ to ``push'' the adoption of the new language standard \aspcoretwo\footnote{\url{https://www.mat.unical.it/aspcomp2013/ASPStandardization/}}
: in 2013 it was not fully supported by most participants, and/or the submitters did not have enough time to integrate new language
features in a completely satisfactory way.

In this paper we report about the design of the Fifth ASP Competition\footnote{\url{https://www.mat.unical.it/aspcomp2014/}}, jointly organized by the University of Calabria (Italy), the Aalto University (Finland), and the University of Genova (Italy). The event is part of the FLoC Olympic Games\footnote{\url{http://vsl2014.at/olympics/}} at VSL 2014, and is affiliated with the 30th International Conference on Logic Programming (ICLP 2014)\footnote{\url{http://users.ugent.be/~tschrijv/ICLP2014/}}. The basis is a re-run of the \solver Track of the 2013 edition, i.e., same ASP solvers, and essentially same domains. Additionally, $(i)$ we have given the participants the opportunity to submit updated versions of the solvers, yet the call was open to new participants; $(ii)$ we have also reconsidered problem encodings, providing new versions for almost all problems; $(iii)$ the domains have not been classified simply by taking into account the ``complexity'' of the encoded problems (as in past events), but also considering the language features involved in the encoding (e.g., choice rules, aggregates, presence of queries). This was intended not only to widen participation, but also in order to be able to measure both the progress of each solver, and of the state of the art in ASP solving, and draw a more complete picture about the approaches that can efficiently solve problems with different features, while still contributing to push the new standard.

The present report is structured as follows. Section~\ref{sec:scoring} illustrates the setting of the competition, while Section~\ref{sec:benchmarks} and Section~\ref{sec:participants} present the problem domains and ASP solvers taking part in the competition, respectively. The report ends by drawing some final remarks in Section~\ref{sec:conclusions}.

\section{Format of the Fifth ASP Competition}\label{sec:scoring}

In this section, we illustrate the settings of the competition, focusing on the differences introduced with respect to the previous editions. The competition 
builds on the basis of the last competition, with each participant in the last edition having the chance (but not the obligation) to submit an updated version, and the focus is on the System Track. 
As already discussed in Section~\ref{sec:introduction}, this can be seen, in a sense, as a special edition of the competition series, and the format changes accordingly, especially when looking at track structure and scoring system.

We decided first to simplify the scoring system w.r.t.\ the last edition (see \cite{4aspcomp}), and to improve it in the case of 
optimization problems. 

As far as the language is concerned, in order to encourage new teams and research groups to join the event, we completely redesigned the tracks, which are now conceived on  language features, other than on a complexity basis. It now makes sense, for a brand new system, or even a preliminary/experimental new version of an old one, to take part only in some tracks, i.e., the ones featuring the subset of the language it correctly supports. In addition, this choice should pave the way to more interesting analyses, such as measuring the progress of a solver, as well of the state of the art, while still contributing to push the standardization process forward, one of the main goal of the competition series. Furthermore, 
the tracks are supposed to draw a clearer and more complete picture about what (combinations of) techniques work for a particular set of features, which, in our opinion, is more interesting, from a scientific point of view, than merely reporting the winners.

In the following, we briefly recall the previous competition format, before discussing the new one: we will illustrate categories and tracks, and present the scoring system in detail, along with a proper overview of general rules; eventually, we will provide the reader with some information about the competition infrastructure.


\paragraph{Previous Competition format.}\label{subsec:previousCompetitionsFormat} The Fourth ASP Competition \cite{4aspcomp} consisted of two different tracks, adopting the distinction between \team and \solver Track. Both tracks featured a selected suite of domains, chosen by means of an open {\em Call for Problems} stage. 
The \solver Track was conceived with the aim of fostering the language standardization, and let the competitors compare each other on fixed encodings under predefined conditions, excluding, e.g., domain-tailored evaluation heuristics and custom problem encodings. The \team Track was instead left open to any system loosely based on a declarative specification language, with no constraints on the declarative language. Indeed, the spirit of this track was to encourage the development of new declarative constructs and/or new modeling paradigms and to foster the exchange of ideas between communities in close relationships with ASP, besides the stimulation of the development of new ad-hoc solving methods, refined problem specifications and solving heuristics, on a per domain basis.

\paragraph{New Competition format.}\label{subsec:newCompetitionsFormat} First of all, given what has already been discussed above, along with the fact that we aim to 
focus on language features, there is no 
\team Track in this edition. 
The competition will take place, then, in the spirit of the former 
\solver Track: it was open to any general-purpose solving system, provided it was able to parse the \aspcoretwo input format. 
Encodings for each problem selected for the competition, along with instance data, have been chosen by the Organizing Committee (see Section~\ref{sec:benchmarks}); participant systems will be run in a uniform setting, on each problem and instance thereof ({\em out-of-the-shelf} setting). Furthermore, sub-tracks are not based on ``complexity'' of problems (as in past events), but rather take into consideration language features, as already discussed above.

\paragraph{Competition Categories.}\label{subsubsec:categories}
The competition consists of {\em two categories}, depending on the computational resources allocated to each running system:
\begin{compactitem}
\item
    {\bf SP}: One processor allowed;
\item
    {\bf MP}: Multiple processors allowed.
\end{compactitem}
While the {\bf SP} category aims at sequential solving systems,
parallelism can be exploited in the {\bf MP} category.

\paragraph{Competition Tracks.}\label{subsubsec:tracks}
As stated by the Call for Participation, according to the availability of benchmarks, to submitted systems, and to participants feedback, both categories of the competition are structured into {\em four tracks}, which are described next. 

\begin{compactitem}
\item
    \textbf{Track \#1}:\ {\em Basic Decision}. Encodings: normal logic programs, simple arithmetic and comparison operators.
\item
    \textbf{Track \#2}:\ {\em Advanced Decision}. Encodings: full language, with queries, excepting optimization statements and non-HCF disjunction.
\item
    \textbf{Track \#3}:\ {\em Optimization}. Encodings: full language with optimization statements, excepting non-HCF disjunction.
\item
    \textbf{Track \#4}:\ {\em Unrestricted}. Encodings: full language.
\end{compactitem}

\paragraph{Scoring system.}\label{subsec:scoring}
The scoring system adopted simplifies the ones from the Third and Fourth ASP Competitions. In particular, it balances the following factors:
\begin{itemize}
\item
    Problems are always weighted equally. 
\item
    If a system outputs an incorrect answer to some instance of a problem, this should invalidate its score for the problem, even if all other instances are correctly solved.
\item
    In case of Optimization problems, scoring is mainly be based on solution quality.
\end{itemize}
In general, 100 points can be earned for each benchmark problem. The final
score of a solving system will hence consist of the sum of scores over all
problems.

\paragraph{Scoring Details.}\label{subsubsec:scoringDetails}
For {\em Decision and Query problems}, the score of a solver $S$ on a problem $P$ featuring $N$ instances is computed as
\[
S(P) = \frac{N_S * 100}{N}
\]
where $N_S$ is the number of instances solved within the allotted time and memory limits.

For {\em Optimization problems}, solvers are ranked by solution quality. Let $M$ be the number of participant systems; then, the score of a solver $S$ for an instance $I$ of a problem $P$ featuring $N$ instances is computed as
\[
    S(P,I) = \frac{M_S(I) * 100}{M * N}
\]
where $M_S(I)$ is
\begin{itemize}
\item
    $0$, if $S$ did neither provide a solution, nor report unsatisfiability, or
\item
    the number of participant solvers that did not provide any strictly better solution than $S$, where a confirmed optimum solution is considered strictly better than an unconfirmed one, otherwise.
\end{itemize}
The score $S(P)$ of a solver $S$ for problem $P$ consists of the sum of scores $S(P,I)$ over all $N$ instances $I$ featured by $P$. Note that, as with Decision and Query problems, $S(P)$ can range from 0 to 100.

\paragraph{Global Ranking.}\label{subsubsec:globalRanking}
The global ranking for each track, and the overall ranking, is obtained by awarding each participant system the sum of its scores over all problems; systems are ranked by their sums, in decreasing order. In case of a draw over sums of scores, the sum of run-times is taken into account as a tie-breaker, in favor, of course, of the system whose run-time is smaller.

\paragraph{Verification of Answers.}\label{subsubsec:detectionOfIncorrectAnswers}
Each benchmark domain $P$ is equipped with a checker program $C_P$ that takes as input both an instance $I$ and a corresponding witness solution $A$, and it is such that $C_P(A,I) ={}${\em true} 
in case $A$ is a valid witness for $I$ w.r.t.\ $P$.

Let us suppose that a system $S$ is faulty for an instance $I$ of a problem $P$; then, there are two possible ways to detect incorrect behavior, and subsequently disqualify system $S$ for $P$:
\begin{itemize}
\item
    $S$ produces an answer $A$, but $A$ is not a correct solution for $I$. This case is detected by checking the outcome of $C_P(A,I)$.
\item
    $S$ recognizes instance $I$ as unsatisfiable, but $I$ actually has some witness solution. In this case, it is checked whether another system $S'$ produced a solution $A'$ for which $C_P(A',I)$ is true.
\end{itemize}

A case of general failure (e.g., ``out of memory'' errors or some other
abrupt system failures) does not imply disqualification on a given benchmark.

When dealing with Optimization problems, checkers produce also the cost of the (last) witness. This latter value is considered when computing scores and assessing answers of systems. 
Given an instance $I$ for an Optimization problem $P$, in general, the cost of best witnesses found over all participants is taken as the {\em imperfect optimum}. When a system $S$ marks its witness $A$ as optimal for $I$:
\begin{itemize}
\item
    if no other system finds a better witness for $I$, $A$ is pragmatically assumed to be optimal;
\item
    if the cost of $A$ turns out to be different from the {\em imperfect optimum} for $I$, $S$ is disqualified on $P$.
\end{itemize}

\paragraph{Software and Hardware settings.}\label{subsec:softwareHardware}
The competition is run on a Debian Linux server (64bit kernel), featuring Intel Xeon X5365 Processors with $8$MB of cache and $16$GB of RAM. Time and memory for each run are limited to 10 minutes and $6$GB, respectively.
Participants can exploit up to $8$ cores in the {\bf MP} category, whereas the execution is constrained to $1$ core in the {\bf SP} category. The execution environment is composed of a number of 
scripts, and performance is measured using the \textit{pyrunlim} tool%
\footnote{\url{https://github.com/alviano/python/}}.



\section{Benchmark Suite}\label{sec:benchmarks}

The benchmark domains in this edition of the ASP Competition largely coincide with the ones from 2013.
Although \aspcoretwo\ encodings had already been made available one year ago,
most participants then lacked preparation time and could not submit appropriate systems.
In fact, half of the systems in 2013 were run on ``equivalent'' encoding reformulations in legacy formats.
In the meantime, however,
the \aspcoretwo\ compliant grounder \Gringof\ became available,
furnishing an off-the-shelf front-end for solvers operating at the propositional level.

As described in Section~\ref{sec:scoring}, the benchmarks in the Fifth ASP Competition are categorized into tracks based on the language features utilized by encodings. Table~\ref{table:problem-list} provides a respective overview, grouping benchmark domains in terms of language features in the \aspcoretwo\ encodings from 2013. That is, the 2013 encodings for \problem{Labyrinth} and \problem{Stable Marriage} belong to the Basic Decision track, and the ``\Dec'' entries in the second column
indicate that both domains deal with Decision problems.
The Advanced Decision track includes the sixteen 2013 encodings for the domains in rows from \problem{Bottle Filling} to   \problem{Weighted-Sequence Problem}. Among them, the \problem{Reachability} domain aims at Query answering, as indicated by ``\Que'' in the second column.
The following four rows marked with ``\Opt'' provide the domains in the Optimization track. Finally, the last four rows give the encodings in the Unrestricted track, where \problem{Abstract Dialectical Frameworks} is an Optimization problem and \problem{Strategic Companies} deals with Query answering.
\begin{table}[t]
  \caption{\label{table:problem-list}Benchmark Suite of the Fifth ASP Competition}%
  \begin{tabular}{|l@{\:}|@{\!\!\!}c@{\:}||@{\!\!}l|@{\!\!}l@{\:}|}
  \cline{1-4}
  \textbf{Domain} & \textbf{P} & \textbf{2013 Encoding} & \textbf{2014 Encoding}
  \\\cline{1-4}\multicolumn{4}{c}{}\gap\cline{1-4}
  \problem{Labyrinth} & \Dec & \ba, \pr & \ba, \pr
  \\\cline{1-4}
  \problem{Stable Marriage} & \Dec & \ba & \ba
  \\\cline{1-4}\multicolumn{4}{c}{}\gap\cline{1-4}
  \problem{Bottle Filling} & \Dec & \ag & \ag, \ch
  \\\cline{1-4}
  \problem{Graceful Graphs} & \Dec & \chc & \chc
  \\\cline{1-4}
  \problem{Graph Colouring}\bas & \Dec & \di & \ba
  \\\cline{1-4}
  \problem{Hanoi Tower}\bas & \Dec & \di & \ba
  \\\cline{1-4}
  \problem{Incremental Scheduling} & \Dec & \ag, \chc & \ag, \chc
  \\\cline{1-4}
  \problem{Knight Tour with Holes}\bas & \Dec & \di, \pr & \ba, \pr
  \\\cline{1-4}
  \problem{Nomystery} & \Dec & \ag, \chc & \chc
  \\\cline{1-4}
  \problem{Partner Units} & \Dec & \ag, \di, \pr & \ag, \ch
  \\\cline{1-4}
  \problem{Permutation Pattern Matching} & \Dec & \chc & \ch
  \\\cline{1-4}
  \problem{Qualitative Spatial Reasoning} & \Dec & \chc, \di & \di
  \\\cline{1-4}
  \problem{Reachability} & \Que & \pr & \out
  \\\cline{1-4}
  \problem{Ricochet Robots} & \Dec & \chc & \ag, \chc
  \\\cline{1-4}
  \problem{Sokoban} & \Dec & \ag, \chc & \chc
  \\\cline{1-4}
  \problem{Solitaire} & \Dec & \chc & \ag, \chc
  \\\cline{1-4}
  \problem{Visit-all}\bas & \Dec & \ag, \chc & \ba
  \\\cline{1-4}
  \problem{Weighted-Sequence Problem} & \Dec & \chc & \ag, \ch
  \\\cline{1-4}\multicolumn{4}{c}{}\gap\cline{1-4}
  \problem{Connected Still Life} & \Opt & \ag, \chc, \pr & \ag, \ch, \pr
  \\\cline{1-4}
  \problem{Crossing Minimization} & \Opt & \di & \ag, \ch
  \\\cline{1-4}
  \problem{Maximal Clique} & \Opt & \di & \ba
  \\\cline{1-4}
  \problem{Valves Location} & \Opt & \ag, \chc, \pr & \ag, \chc, \pr
  \\\cline{1-4}\multicolumn{4}{c}{}\gap\cline{1-4}
  \problem{Abstract Dialectical Frameworks} & \Opt & \ag, \di, \lev, \pr & \ag, \di, \lev, \pr
  \\\cline{1-4}
  \problem{Complex Optimization} & \Dec & \ch, \di, \pr & \ch, \di, \pr
  \\\cline{1-4}
  \problem{Minimal Diagnosis} & \Dec & \di, \pr & \di, \pr
  \\\cline{1-4}
  \problem{Strategic Companies} & \Que & \di, \pr & \out
  \\\cline{1-4}
  \end{tabular}
\end{table}

The third column of Table~\ref{table:problem-list} indicates particular language features
of the encodings from the Fourth ASP Competition.
While merely normal rules and comparison operators,
considered as ``\ba'' features, are used for \problem{Stable Marriage}, 
the Basic Decision encoding for \problem{Labyrinth} induces ``\pr'' ground instances
with positive recursion among atoms
\cite{fage-94,erde-lifs-2003}.
The use of aggregates like \countagg, \sumagg, \maxagg, and \minagg\
\cite{fabe-etal-2008-tplp,asp-core-2-01c}, e.g.,
in the Advanced Decision encoding for \problem{Bottle Filling},
is indicated by an ``\ag'' entry.
Moreover, ``\di'' denotes proper disjunctions in rule heads
\cite{gelf-lifs-91,eite-gott-95},
as utilized in the 2013 encoding for \problem{Graph Colouring}.
Choice rules \cite{simo-etal-2002,asp-core-2-01c}
are indicated by a ``\ch'' entry, where the superscript ``\chb''
stands for non-trivial lower and/or upper bounds on the number
of chosen atoms, e.g.,
used in the Advanced Decision encoding for \problem{Graceful Graphs}.
Unlike that, the choice rules for \problem{Complex Optimization}
in the Unrestricted track are unbounded, and thus ``\chb'' is omitted in its row.
Finally, \problem{Abstract Dialectical Frameworks} is the only
Optimization problem in the Fifth ASP Competition
for which more than one ``\lev''
\cite{simo-etal-2002,leon-etal-2002-dlv}
is used in the encoding.

Compared to the Fourth ASP Competition,
we decided to drop the \problem{Chemical Classification} domain,
whose large encoding imposed primarily a grounding bottleneck.
On the other hand, we reintroduced the application-oriented
\problem{Partner Units} domain, reusing encodings and instances
submitted to the Third ASP Competition.
In order to furnish a novel benchmark collection for this year,
we also devised new encoding variants utilizing the language features
indicated in the fourth column of Table~\ref{table:problem-list}.
The 2014 encodings for the domains marked with~``\bas'' omit
advanced language features of their 2013 counterparts, so that the
Basic Decision track on new encodings comprises six benchmark domains.
By evaluating the participant systems on previous as well as new encodings,
we hope to gain insights regarding the impact of encodings on system performance.
In fact, for all domains but \problem{Reachability} and
\problem{Strategic Companies} aiming at Query answering,
we could make substantial modifications to previously available encodings.

The instances to run in the Fifth ASP Competition have been randomly selected from the suites submitted in 2013 (or 2011 for \problem{Partner Units}), using the concatenation of winning numbers from the EuroMillions lottery of Tuesday, 22nd April 2014, as random seed. In this way, twenty instances were picked per domain in order to assess the participant systems both on the encodings from 2013 as well as their new variants.

\section{Participants}\label{sec:participants}
In this section, we briefly present all participants; we refer the reader to the official competition website \cite{aspcomp2014-web} for further details.

The competition featured 16 systems coming from three teams:


\begin{compactitem}
\item
    The Aalto team from the Aalto University submitted nine solvers, 
    working by means of translations
    \cite{boma-13,gebs-etal-2014,ljn-kr-12,DBLP:journals/corr/abs-1108-5837}.
    Three systems,
    {\sc {\small lp2sat3+glucose}}, {\sc {\small lp2sat3+lingeling}}, and {\sc {\small lp2sat3+plingeling-mt}},
    rely on translation to SAT,
    which includes the normalization of aggregates as well as
    the encoding of level mappings for non-tight problem instances.
    The latter part is expressed in terms of bit-vector logic
    or acyclicity checking, respectively, supported by the back-end SMT
    solvers of the {\sc {\small lp2bv2+boolector}} and {\sc {\small lp2graph}} systems.
    While the aforementioned systems do not support optimization and participate in the
    Basic and Advanced Decision tracks (\#1 and \#2) only,
    {\sc {\small lp2maxsat+clasp}}, {\sc {\small lp2mip2}}, and {\sc {\small lp2mip2-mt}},
    running \clasp\ as a Max-SAT solver or the Mixed Integer Programming solver {\sc {\small cplex}}
    as back-ends, respectively, compete in the Optimization track (\#3) as well.
    Finally, {\sc {\small lp2normal2+clasp}} normalizes aggregates (of up to certain size)
    and uses \clasp\ as back-end ASP solver;
    {\sc {\small lp2normal2+clasp}} participates in all four tracks and
    thus also in the Unrestricted track (\#4).
    All systems by the Aalto team utilize \Gringof\ for grounding, and neither of them supports
    Query problems (\problem{Reachability} and \problem{Strategic Companies}).
    The systems {\sc {\small lp2sat3+plingeling-mt}} and {\sc {\small lp2mip2-mt}}
    exploit multi-threading and run in the {\bf MP} category,
    while the other, sequential systems participate in the {\bf SP} category.

\item
    The Potassco team from the University of Potsdam submitted the sequential system
    \clasp\ \cite{gks2012-aij} in the {\bf SP} category and the multi-threaded
    \claspmt\ system \cite{gekasc12b} in the {\bf MP} category.
    Both systems utilize \Gringof\ for grounding and \clasp,
    a native ASP solver for (extended) disjunctive logic programs
    based on conflict-driven learning,
    as back-end.
    The systems participate in all tracks, except for the Query problems.

\item 
The Wasp team from the University of Calabria submitted five incarnations of
\wasp\ \cite{wasp-lpnmr-13,alvi-etal-2014-nmr},
a native ASP solver built upon 
techniques originally introduced in SAT, yet 
extended and 
combined with techniques specifically designed for solving disjunctive logic programs,
in the {\bf SP} category.
Unlike {\sc {\small wasp-1}}, utilizing a prototype version of {\sc {\small DLV}} (to cope with the \aspcoretwo\ language) for grounding,
\mbox{{\sc {\small wasp-2}}} relies on \Gringof\ 
and further differs from {\sc {\small wasp-1}} in the implementation of support inferences and program simplifications.
Moreover, {\sc {\small wasp-1.5}} is a hybrid system combining
{\sc {\small wasp-1}} and {\sc {\small wasp-2}}, basically switching
between them depending on whether a logic program is HCF or subject to a Query.
While {\sc {\small wasp-1}} and {\sc {\small wasp-1.5}} compete in all domains and tracks,
{\sc {\small wasp-2}} does not participate in the Unrestricted track (\#4).
Finally, the {\sc {\small wasp-wmsu1-only-weak}} and {\sc {\small wasp-wpm1-only-weak}} systems
are specifically designed for solving optimization problems and thus participate
in the Optimization track (\#3) only.
%
%
\end{compactitem}

In summary, similarly to past competitions, we can identify two main
ASP solving approaches:
\begin{compactitem}
\item
``native'' systems, which exploit techniques purposely
conceived/adapted for dealing with logic programs under the stable
models semantics, and
\item
``translation-based'' systems, which (roughly)
at some stage of the evaluation process produce an intermediate specification
in some different formalism, which is then fed to a corresponding solver.
 \end{compactitem}
The solvers submitted by the Potassco and Wasp teams 
as well as the {\sc {\small lp2normal2+clasp}} system by the Aalto team
rely on the first approach,
while the remaining systems 
by the Aalto team are of the second kind.

It is worth mentioning that, in order to assess the improvements in system implementation,
we also ran a selection of systems that were submitted in the Fourth ASP Competition.
In particular, we considered one system per team selected according to the following criteria:
$(i)$ it features an updated version in this year's edition, 
$(ii)$ it is compliant with \aspcoretwo, and
$(iii)$ it 
performed best in the Fourth ASP Competition among the systems submitted by the same team.



\section{Conclusions}\label{sec:conclusions}
The Fifth ASP Competition is jointly organized by the University of Calabria (Italy), the Aalto University (Finland), and the University of Genova (Italy), and is affiliated with the 30th International Conference on Logic Programming (ICLP 2014). The main goals of the Fifth ASP Competition are to measure the advances of the state of the art in ASP solving  and to push the adoption of the \aspcoretwo standard format.
In this paper, the design of the fifth edition of the ASP Competition was presented, along with an overview of the participants.

The results will be announced in Vienna at ICLP 2014, which is part of the Federated Logic Conference. Participants will be awarded in a ceremony organized by the FLoC Olympic Games at the Vienna Summer of Logic, taking place on Monday, 21th July 2014.


\newif\ifmakebbl
\makebblfalse
\ifmakebbl
\bibliographystyle{acmtrans} 
\bibliography{bibtex}\label{sec:bib}
\else
\newcommand{\SortNoOp}[1]{}

\fi

\end{document}